\documentclass{article} % For LaTeX2e
\usepackage{times}
\usepackage[accepted]{icml2025}

\usepackage{amsmath,amsfonts,bm}

\def\eqref#1{equation~\ref{#1}}
\def\1{\bm{1}}

\DeclareMathAlphabet{\mathsfit}{\encodingdefault}{\sfdefault}{m}{sl}
\SetMathAlphabet{\mathsfit}{bold}{\encodingdefault}{\sfdefault}{bx}{n}

\usepackage{color}
\definecolor{reb}{RGB}{0,0,255}
\usepackage{hyperref}
\usepackage{url}

\usepackage{multirow}
\usepackage{graphicx}
\usepackage{booktabs}
\usepackage{color}
\usepackage{colortbl}
\usepackage{cleveref}
\usepackage{wrapfig}
\usepackage{tcolorbox}
\usepackage{pifont}

\definecolor{my_green}{RGB}{51,102,0}
\definecolor{my_yellow}{RGB}{255,165,0}
\definecolor{my_red}{RGB}{204, 0, 0}
\newcommand{\colorcmark}{\textcolor{my_green}{\ding{52}}}
\newcommand{\colorxmark}{\textcolor{my_red}{\ding{55}}}

\begin{document}

\twocolumn[
\icmltitle{LoRA-Gen: Specializing Large Language Model via Online LoRA Generation}



\icmlsetsymbol{equal}{*}
\icmlsetsymbol{corr}{$\dagger$}

\begin{icmlauthorlist}
\icmlauthor{Yicheng Xiao}{thu,equal}
\icmlauthor{Lin Song}{comp,equal}
\icmlauthor{Rui Yang}{hku}
\icmlauthor{Cheng Cheng}{xjtu}
\icmlauthor{Yixiao Ge}{comp}
\icmlauthor{Xiu Li}{thu,corr}
\icmlauthor{Ying Shan}{comp}
\end{icmlauthorlist}

\icmlaffiliation{thu}{Tsinghua University}
\icmlaffiliation{comp}{ARC Lab, Tencent PCG}
\icmlaffiliation{hku}{The University of Hong Kong}
\icmlaffiliation{xjtu}{Xi’an JiaoTong University}

\icmlcorrespondingauthor{Xiu Li}{li.xiu@sz.tsinghua.edu.cn}

\icmlkeywords{Machine Learning, ICML}

\vskip 0.3in
]



\printAffiliationsAndNotice{\icmlEqualContribution} % otherwise use the standard text.

% \title{LoRA-Gen: Specializing Language Model via Online LoRA Generation}

% % Authors must not appear in the submitted version. They should be hidden
% % as long as the \iclrfinalcopy macro remains commented out below.
% % Non-anonymous submissions will be rejected without review.

% \author{Antiquus S.~Hippocampus, Natalia Cerebro \& Amelie P. Amygdale \thanks{ Use footnote for providing further information
% about author (webpage, alternative address)---\emph{not} for acknowledging
% funding agencies.  Funding acknowledgements go at the end of the paper.} \\
% Department of Computer Science\\
% Cranberry-Lemon University\\
% Pittsburgh, PA 15213, USA \\
% \texttt{\{hippo,brain,jen\}@cs.cranberry-lemon.edu} \\
% \And
% Ji Q. Ren \& Yevgeny LeNet \\
% Department of Computational Neuroscience \\
% University of the Witwatersrand \\
% Joburg, South Africa \\
% \texttt{\{robot,net\}@wits.ac.za} \\
% \AND
% Coauthor \\
% Affiliation \\
% Address \\
% \texttt{email}
% }

% The \author macro works with any number of authors. There are two commands
% used to separate the names and addresses of multiple authors: \And and \AND.
%
% Using \And between authors leaves it to \LaTeX{} to determine where to break
% the lines. Using \AND forces a linebreak at that point. So, if \LaTeX{}
% puts 3 of 4 authors names on the first line, and the last on the second
% line, try using \AND instead of \And before the third author name.

\newcommand{\fix}{\marginpar{FIX}}
\newcommand{\new}{\marginpar{NEW}}

% %\iclrfinalcopy % Uncomment for camera-ready version, but NOT for submission.
% \begin{document}

% \maketitle

\begin{abstract}
Recent advances have highlighted the benefits of scaling language models to enhance performance across a wide range of NLP tasks. However, these approaches still face limitations in effectiveness and efficiency when applied to domain-specific tasks, particularly for small edge-side models.
We propose the LoRA-Gen framework, which utilizes a large cloud-side model to generate LoRA parameters for edge-side models based on task descriptions.
By employing the reparameterization technique, we merge the LoRA parameters into the edge-side model to achieve flexible specialization.
Our method facilitates knowledge transfer between models while significantly improving the inference efficiency of the specialized model by reducing the input context length.
Without specialized training, LoRA-Gen outperforms conventional LoRA fine-tuning, which achieves competitive accuracy and a 2.1x speedup with TinyLLaMA-1.1B in reasoning tasks.
Besides, our method delivers a compression ratio of 10.1x with Gemma-2B on intelligent agent tasks.

\end{abstract}

\section{Introduction}
The principle of scaling laws~\citep{kaplan2020scaling} demonstrates that increasing the size of Large Language Models (LLMs) can significantly improve cross-task generalization.
However, due to the constraints of their enormous size, generic LLMs struggle to achieve a good balance between efficiency and effectiveness when addressing domain-specific tasks or preferences.
Consequently, research has been shifted towards developing more specialized, compact language models optimized for specific tasks and capable of local deployment on edge devices~\citep{fu2023specializing,grangier2024specialized,shen2024tag}. This emerging approach addresses the critical need for more adaptable and resource-efficient AI solutions across academic and industrial domains.
Many approaches utilize parameter-efficient fine-tuning techniques~\citep{peft-1,prefix-tuning, prompt-tuning, lora}, particularly LoRA~\citep{lora}, to train on specific datasets for specialization.
However, this method may encounter the issue of catastrophic forgetting, which can result in a decrease in performance on other unseen tasks~\citep{forget-claim-1,forget-claim-2}.

\begin{figure}[t]
\centering
\includegraphics[width=0.4\textwidth]{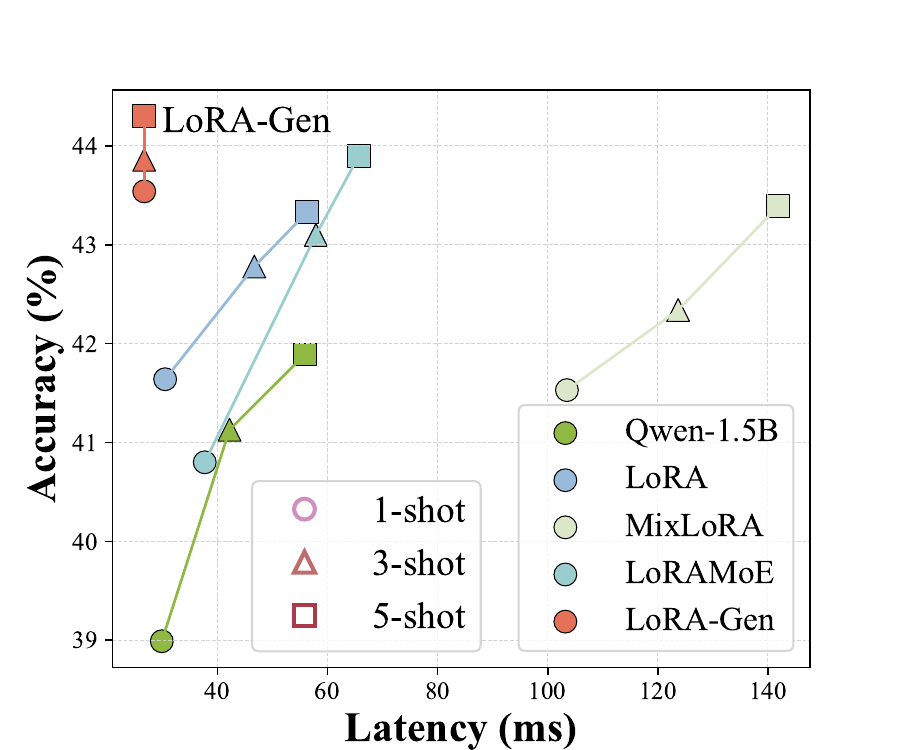}
\caption{\footnotesize Accuracy-latency curves comparison with various few-shot numbers on ARC-c task. Best view in color. Base model is Qwen-1.5B.}
\vspace{-10pt}
\label{fig:curve}
\end{figure}

\begin{figure*}[th]
    \centering
    \includegraphics[width=0.9\linewidth]{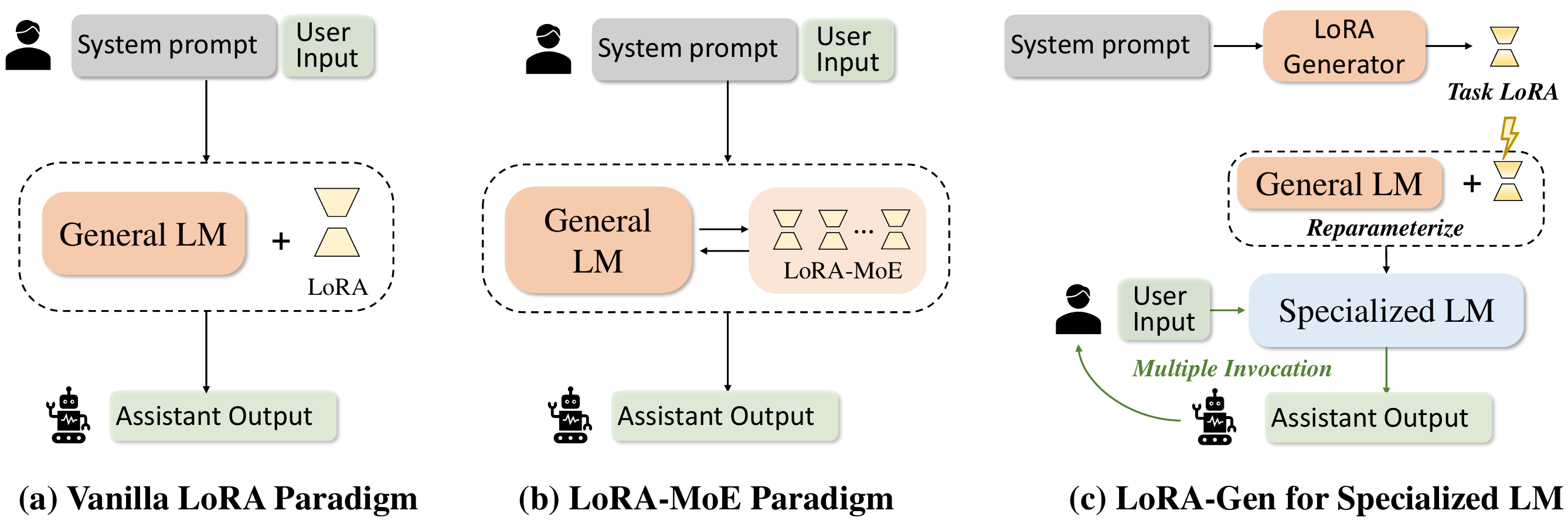}
    \caption{Comparison of different LoRA-based fine-tuning strategies.(a) Vanilla LoRA is fine-tuned on the target task and then merged into the source model. (b) LoRA-MoE introduces additional LoRA experts to improve the generalization performance. (c) Our LoRA-Gen presents a task-specific LoRA generator that customizes a specialized LM for edge-side users.}
    \label{fig:teaser}
\end{figure*}
\begin{table*}[h]
    \centering
    \footnotesize
    \renewcommand\arraystretch{1.0}
    \setlength{\tabcolsep}{3pt}
    \begin{tabular}{l|cccc}
    \toprule
    \multirow{3}{*}{Method} & Context Compression & Reparameterized & Training Free & Knowledge \\
    &for Unseen Tasks & Model & for Unseen Tasks & Transfer \\
    & (Fast Inference) & (w/o Additional Params.) & (High Flexibility) & (High Accuracy) \\
    \midrule
    ICL~\cite{dong2022survey} & \colorxmark & \colorxmark & \colorcmark  & \colorxmark \\
    LoRA~\cite{lora} & \colorxmark & \colorcmark & \colorxmark  & \colorxmark \\
    % MixLoRA &  & \colorcmark &  \\
    LoRA-MoE~\cite{loramoe} & \colorxmark & \colorxmark & \colorcmark  & \colorxmark \\
    % MoLA &  &  &  \\
    LoraHub~\cite{lorahub} & \colorxmark & \colorcmark & \colorxmark  & \colorxmark \\
    AutoCompressors~\cite{ChevalierWAC23} & \colorcmark & \colorxmark & \colorcmark  & \colorxmark \\
    \rowcolor{gray!10} LoRA-Gen & \colorcmark & \colorcmark & \colorcmark  & \colorcmark \\
    \bottomrule
    
    \end{tabular}
    \caption{Characteristics comparison with other counterparts. ICL indicates the in-context learning. }
    \label{tab:introduction}
    % \vspace{-10pt}
\end{table*}

To alleviate knowledge forgetting in specialized training, recent approaches~\citep{loramoe,mola,moral,mixlora}, leverage the flexibility of the Mixture of Experts (MoE) for LoRA training.
Specifically, as shown in~\Cref{fig:teaser}(b), they integrate a group of multiple LoRA components as experts within the language model, allowing the language model to control the selection of LoRA components during token generation.
However, these methods introduce additional inference costs due to the extra experts and control units.
LoRAHub~\citep{lorahub}, on the other hand, pre-trains a set of task-specific LoRA components and employs a manually designed parameter-free optimization method for selection.
Nevertheless, the effectiveness of above mentioned approaches is limited by their model scale, resulting in constrained performance and generalization capabilities on unseen tasks.
Therefore, this paper explores a new perspective: \textit{utilizing a large cloud-side model to generate parameters for a smaller edge-side model to achieve better specialization.}

To achieve it, we propose a new LoRA generation framework, termed LoRA-Gen.
As shown in \Cref{fig:teaser}(c), our method can be divided into two parts: Online LoRA generation and Specialized LM.
The former is used to generate LoRA parameters based on the task-defined system prompt, while the latter facilitates efficient batch inference for user input.
Specifically, a fine-tuned large language model and a mixture of LoRA experts are deployed in the cloud.
The cloud-side language model generates a set of meta tokens based on the given system prompt.
Each meta token corresponds to a transformer layer in the edge-side language model, utilizing these tokens to control the composition of parameters from the LoRA experts.
Similarly to vanilla LoRA, the combined parameters are further merged into the edge-side LM through reparameterization, resulting in an efficient specialized model.

As shown in ~\Cref{tab:introduction}, our LoRA-Gen offers four advantages over previous methods: i) Context compression for unseen tasks: LoRA-Gen dynamically compresses the task-specific system prompt ($e.g.$, task descriptions, few-shot samples, and chat templates) into the LoRA weights, which significantly reduces the context length for the specialized models. ii) Reparameterized model: Unlike LoRA-MoE~\citep{loramoe}, our approach employs reparameterization techniques to merge the generated LoRA weights into the original parameters, thereby avoiding additional inference costs. iii) Training free for unseen tasks: Our method does not require any additional training, including few-shot tuning, when specializing the model for unseen tasks. It only necessitates a single-turn inference on the system prompt to obtain the specialized model parameters, which simplifies model deployment. iv) Knowledge Transfer: LoRA-Gen allows the cloud side and edge side to utilize different models, enabling the injection of knowledge from the large cloud model into the edge model through reparameterization, which enhances performance effectively as shown in~\Cref{fig:curve}.

We conduct extensive experiments to validate the effectiveness of LoRA-Gen on various commonsense reasoning tasks as well as an agent benchmark.
The results demonstrate that our method balances both performance and efficiency, showing significant advantages across eight language datasets.
For the edge-side model of TinyLLaMA-1.1B, LoRA-Gen outperforms vanilla LoRA fine-tuning by a remarkable margin with only $16\%$ sequence length, $+$1.3\% on harmonic-mean of accuracy, and 2.1x speedup.
Moreover, for the Gemma-2B model, LoRA-Gen demonstrates competitive performance on unseen agent tasks.
Additionally, since it does not require the input of agent definitions during inference, it achieves a remarkable 10.1x compression ratio.

\section{Related Work}
\label{related}
\subsection{Parameter-Efficient Fine-Tuning}

Given the billions of parameters in LLMs and the limitations of current hardware, fully fine-tuning LLMs in the traditional manner is often impractical. To address this, several parameter-efficient fine-tuning (PEFT) methods have been developed. Adapter-based approaches~\cite{MahabadiHR21, zhou2024uniqa, ZhangHLZL00024} involve inserting trainable adapter layers into various blocks of pre-trained models. Soft prompt methods~\cite{prefix-tuning,LiuTMMHBR22} adjust a small trainable prefix vector to adapt LLMs to new tasks. Unlike these methods, LoRA~\cite{lora} minimizes the number of trainable parameters for downstream tasks by freezing the pre-trained models and tuning only additional rank decomposition layers. This method approximates weight adjustments during fine-tuning without incurring extra costs during inference. Building on this, AdaLoRA~\cite{zhang2023adalora} dynamically adjusts the parameter budget among weight matrices, while DoRA~\cite{dora} fine-tunes both the magnitude and directional components decomposed from pre-trained weights. VeRA~\cite{VeRA} further reduces the number of trainable parameters by utilizing shared low-rank layers and learnable scaling vectors.

\subsection{LoRA Meets Mixture of Experts}

Leveraging its lightweight nature, LoRA is utilized in Mixture of Experts (MoE) architectures to enhance performance. MoLoRA~\cite{zadouri2023pushing} incorporates LoRA adapters as experts on top of pre-trained models and uses a router layer to integrate these experts. MOELoRA~\cite{Liu00ZX0024} applies this framework to various medical domain tasks, though it requires task type input for the router. LoRAMoE~\cite{loramoe} introduces multiple LoRA experts into the feed-forward block to mitigate knowledge forgetting during the instruction-tuning phase. LoraHub~\cite{lorahub} allows a dynamic assembling of LoRA modules on various tasks and even unseen tasks by combining adapted LoRA modules. Additionally, MoLA~\cite{mola} proposes layer-specific experts, allocating a varying number of LoRA experts to different layers to boost performance.

\subsection{Context Compression}

With the rise of in-context learning~\cite{CoT} and agentic pipelines~\cite{gpt4tools}, LLMs often need to process thousands of tokens, potentially exceeding their maximum context length. Unlike methods that extend the context window of LLMs, context compression offers an efficient way to reduce the input prompt length. There are two primary methods of context compression: hard prompt and soft prompt. Selective-Context~\cite{li2023unlocking} and \citet{jiang2023llmlingua} exemplify hard prompt methods by removing low-information content at the lexical level (e.g., sentences, words, or tokens) to shorten the prompt. On the other hand, gisting~\cite{Mu0G23}, AutoCompressors~\cite{ChevalierWAC23}, ICAE~\cite{00010WWCW24}, and 500xCompressor~\cite{li2024500xcompressor} represent soft prompt methods that compress input prompts into a small number of special tokens. In contrast to these approaches, we propose compressing the context into rank-decomposition layers using LoRA methods.

\section{Methodology}
\label{method}

In this section, we first review LoRA-based Mixture of Experts fine-tuning paradigm and then elaborate on our LoRA-Gen, which generates task-specific LoRA weights according to the system prompt for edge-side language models.
\subsection{Revisiting Mixture of LoRA Experts}
LoRA~\citep{lora} improves the efficiency of fine-tuning by significantly reducing the number of trainable parameters.
Formally, it updates the weight matrix $W\in \mathbb{R}^{d^\prime \times d^{\prime\prime}}$ by using a low-rank approximation via two decomposition matrices $A \in \mathbb{R}^{d^\prime\times r}$ and $B \in \mathbb{R}^{r\times d^{\prime\prime}}$ with a low rank $r$ ($r\ll min(d^\prime,d^{\prime\prime})$) as follow: 
\begin{equation}
    \widetilde{W} = W + AB.
\end{equation}
Trainable low-rank decomposition matrices can capture the underlying patterns of downstream tasks under the guidance of the task-specific direction~\citep{lora}.
Moreover, another effective approach, the Mixture of Experts (MoE)~\citep{MoE,hierarchical_moe}, treats multiple networks as experts and seeks to take advantage of their strengths in a hybrid framework. This method aims to combine the advantages of different models, resulting in improved generalization and overall performance.
Typically, a MoE layer consists of $n$ experts, denoted as $\{E_i\}_{i=1}^n$ with a router $R$ as the gate for expert allocation.
Given hidden states $\{h_j\}_{j=1}^s$ of a sequence with the length of $s$, the output of the MoE can be formulated as:
\begin{equation}
    h^{\prime}_{j} = {\sum_{i=1}^{n}} R_i(h_j)E_i(h_j) 
\end{equation}
Considering the efficiency of LoRA and the strong performance of MoE,~\cite{mixlora,loramoe,mola,moral} integrate LoRA into the MoE plugin, boosting the fine-tuning performance by utilizing a mixture of LoRA experts, effectively blending the strengths of both methods.
\subsection{Online LoRA Generation}
\paragraph{Overview.}
The mixture of LoRA experts has showcased reasonable performance in fine-tuning for specific tasks.
However, there remains a gap in its effectiveness for multi-task learning and the generalization to unseen tasks.
Additionally, most LoRA-MoE~\citep{mixlora, loramoe} methods require calculating the expert routing for each token individually, which significantly increases the computational complexity.
\begin{figure*}
    \centering
    \includegraphics[width=\linewidth]{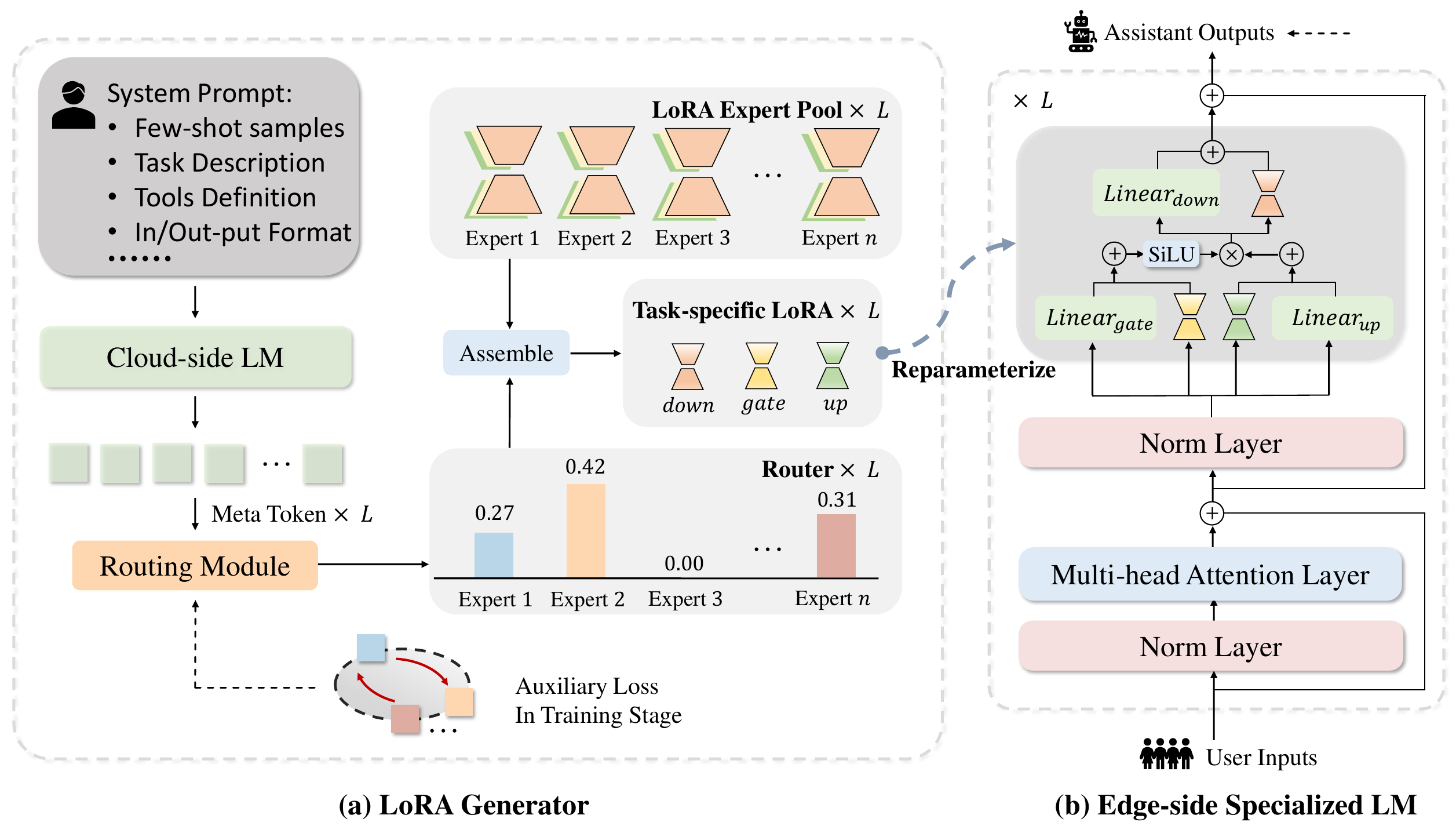}
    \caption{Overview of our proposed LoRA-Gen. Given the system prompts by users, a large language model first generates meta tokens autoregressively. With a routing module, we obtain the gates of all experts in the online LoRA pool. After assembling, we produce the specialized LoRA in the cloud side and deploy it to the edge-side language model by merging the LoRA weights.}
    \label{fig:framework}
\end{figure*}
To address these challenges, we propose a new framework, termed LoRA-Gen that generates task-aware LoRA via an online large language model with system prompts (including few-shot samples, task description, role specification and the conversation format) as presented in~\Cref{fig:framework}.
In the following, we elaborate on our LoRA generation method and the reparameterization of the edge-side language model.

\paragraph{Cloud-side LM \& Meta Token.}
In adherence to meta-learning~\citep{meta-1,meta-2}, we construct a unified representation of the task-related information to achieve generalization capabilities for various tasks, relying on cloud-side LM to facilitate this process. Specifically, given a series of few-shot samples or task-specific system prompts, the cloud-side LM appends $L$ special tokens $\langle meta\rangle$ behind them and transfers the inherent knowledge into these tokens with causal masks in a single forward pass.
We define these tokens as meta tokens $\{T_i^{meta}\}_{i=1}^L$, where $L$ represents the number of layers of the edge-side language model.
Each meta token is associated with a transformer layer in the edge-side LM.

\paragraph{LoRA Expert Pool.} 
Our initial attempt is to generate LoRA parameters directly through a continuous projection on the meta token.
However, the expansive parameter space poses optimization challenges, making the model susceptible to overfitting and hindering generalization, whose analysis refers to \Cref{tab:lora_generate}.
Therefore, similar to the previous works~\citep{loramoe}, we adopt an alternative solution by introducing the discrete MoE mechanism.
Specifically, as shown in ~\Cref{fig:framework}, we construct a LoRA expert pool of $n$ experts, whose weights are defined as $\{E_i\}_{i=1}^{n}$.
Each LoRA expert contains three LoRA blocks, corresponding to the gate linear layer, up linear layer and down linear layer in FFN of the edge-side model, respectively.
Different from the LoRAHub~\citep{lorahub}, these experts are trained in an end-to-end manner.

\paragraph{Routing Module.}
To control the composition of experts, we propose a routing module using meta tokens.
Unlike the token-wise LoRA-MoE~\citep{loramoe}, our MoE is layer-wise.
We apply an individual MoE for each transformer layer in the edge-side LM, and all tokens in a sequence use the same composition.
For simplicity, the routing module consists of two linear projections with a Batch Normalization (BN) layer.
Incorporating a BN layer can further increase the diversity of router output, promoting the utilization of a wider range of experts.
In formal, the router $R^i \in \mathbb{R}^n$ of $i$-th layer of edge-side LM can be formulated as:
\begin{equation}
    R^i = \text{BN}(f_2\circ \varsigma \circ f_1(T_i^{meta})),
\end{equation}
where $f_1$, $f_2$ are the linear transform and $\varsigma$ denotes the $\text{SiLU}$~\cite{elfwing2018sigmoid} activation function.
We attempt to increase selection randomness and balance expert loads, by using Gumbel-Softmax~\citep{gumbel-softmax}, which can be formulated as:
\begin{align}
\text{Gumbel-Softmax}({R_{t}^{i}})=\frac{e^{R_{t}^{i}+g}}{\sum_{j=1}^{n} e^{R_{j}^{i}+g}}, \\
\text{where}~g\sim\text{Gumbel}(0,1).
\end{align}
Nevertheless, the Gumbel-softmax strategy shows a significant reduction in generalization performance, which is reported in experiments of~\Cref{sec:ablation},
To this end, following~\cite{mixlora,loramoe}, we adopt a KeepTOP-K strategy to select experts in a deterministic manner:
\begin{align}
G_t^i = \left\{\begin{matrix}
  \frac{\widetilde{R_t^i}}{ {\textstyle \sum_{j=1}^{K} \overline{R_j^i}} } & \widetilde{R_t^i} \in \text{TOP-K}(\widetilde{R^i}) \\
  0& \text{else}
\end{matrix}\right. , \\
\text{where}~{\text{TOP-K}(\widetilde{R^i})=\{\overline{R_t^i}\}_{t=1}^{K} }, \widetilde{R_t^i} = \frac{e^{R^i_t}}{ {\textstyle \textstyle \sum_{j=1}^{n} e^{R_j^i}} },
\end{align}
where $G_t^i$ represents the the gate score of $t$-th experts for $i$-th decoder layer of the edge-side language model.
Consequently, we generate task-specific LoRA weights as:
\begin{equation}
\theta^i = \sum_{j=1}^{n}G  ^iE_j .
\end{equation}
where the $\theta^i$ indicates the generated LoRA weights for $i$-th decoder layer.
\paragraph{Reparametrization.}
As the same as LoRA, we use the reparameterization strategy to merge the generated LoRA parameters into the FFN layers of the edge-side model.
In contrast to the LoRA-MoE, our method is cost-free during inference, which needs no additional components in the specialized edge-side LM.

\subsection{Training Target}
\paragraph{Auxiliary Loss.}
Balanced load of MoE structure is essential for capability of generalization and stability~\citep{MoE}.  
Without constraints, the routing module tends to select a fixed small set of experts, leaving other experts unused and causing load imbalance.
To mitigate this issue, we introduce a soft constraint with the coefficient of variation as the auxiliary loss, encouraging a more balanced usage of the available experts.
Formally, the constraint can be formulated as:
\begin{equation}
\mathcal{L}_{cv} = \alpha(\frac{\sigma(G) }{\mu(G) })^2,
\end{equation}
where $\sigma$ and $\mu$ represent the standard deviation and mean of the gates assigned to each expert within a batch, separately.
The coefficient $\alpha$ is to balance the auxiliary objective and the main objective.
\paragraph{Total Loss.}
The total loss is consist of the language modeling loss and auxiliary loss as follows:
\begin{equation}
\mathcal{L}_{total} = \mathcal{L}_{cv} + \mathcal{L}_{LM},
\end{equation}
where $\mathcal{L}_{LM}$ is the Cross Entropy loss of language modeling in causal LMs.

\section{Experiments}
We conduct extensive experiments to evaluate the effectiveness of our LoRA-Gen and compare it to the widely adopted LoRA-based fine-tuning method on commonsense reasoning tasks in a fair experimental setting.
Furthermore, we assess the generalization capacity and system prompt compression performance of LoRA-Gen on an agent dataset, GPT4Tools~\citep{gpt4tools}.
\subsection{Datasets and Metrics.}
\paragraph{Reasoning Tasks.}
Following~\cite{loramoe,mixlora}, we select eight widely-used benchmarks to assess the reasoning ability of LoRA-Gen across various knowledge domains ranging from natural science to daily life.
One classification task: BoolQ~\citep{boolq}.
Five question-answering tasks: ARC-c~\citep{arc}, ARC-e~\cite{arc}, OpenBookQA~\citep{obqa}, PIQA~\citep{piqa} and SocialQA~\citep{siqa}.
One science completion task: Hellaswag~\citep{hellas} and a fill-in-the-blank task: Winogrande~\citep{winog}.
\paragraph{Agent Dataset.}
We utilize the GPT4Tools~\cite{gpt4tools} which provides a benchmark to evaluate the ability of LLM to use tools, to assess the effectiveness of LoRA-Gen in the deployment of intelligent agents.
GPT4Tools constructs a tool-related instructional dataset, including positive samples, negative
samples, and context samples.
It consists of 71k instruction-response pairs with 21 tools in the training set and 652 items in the test set with 8 novel tools absent from the training set.

\paragraph{Metrics.}
The performance of all commonsense reasoning benchmarks is measured with the accuracy metric in all datasets.
To further evaluate the performance in multi-task learning, we utilize two metrics: the average accuracy (AVE.) and the harmonic mean (HAR.) of all results.
For GPT4Tools, we measure the performance of method from five aspect: successful rate of thought ($\mathrm{SR}_t$), successful rate of action ($\mathrm{SR}_{act}$), successful rate of arguments ($\mathrm{SR}_{args}$), successful Rate ($\mathrm{SR}$) and IoU according to~\cite{gpt4tools}.
\begin{table*}[t]
    \centering
    \footnotesize
    \renewcommand\arraystretch{1.0}
    \setlength{\tabcolsep}{2.35pt}
    % \begin{tabular}{l | c | c c c | c c c}
    \begin{tabular}{l|ccccc|ccc|ccc}
    \toprule
    \multirow{2}{*}{Method} & \multicolumn{5}{c}{Seen Tasks}  & \multicolumn{3}{c|}{Unseen Tasks} & \multirow{2}{*}{AVE. $\uparrow$} & \multirow{2}{*}{HAR. $\uparrow$} & {Latency}\\
    \cline{2-9} 
    &  ARC-c & ARC-e & OBQA & BoolQ &SIQA & HellaS & WinoG & PIQA & & &(ms) $\downarrow$ \\ 
    \midrule
    TinyLlaMA-1.1B & 34.2   &66.9 & 27.4 & 58.8 & 46.0 & \textbf{45.8} & 60.7  & 73.9 & 51.7& 46.7 & 44.5\\   
    +LoRA &33.6&67.6&28.6&71.9&51.5&44.5&61.9&\textbf{75.1}&54.3&48.5&44.5 \\
    +LoRAMoE &35.2&68.8&28.6&73.2&\textbf{52.1}&45.4&62.0&74.1&54.9&49.3& 55.9\\   
    +MixLoRA &33.5&67.7&28.4&73.3&51.4 & 44.9& 62.3  & 74.6& 54.5& 48.6 & 100.1\\   
    \rowcolor{gray!10} +LoRA-Gen & \textbf{35.8} &\textbf{69.1} &\textbf{30.4} & \textbf{73.6} & {49.6} & {45.5} & \textbf{62.6} & {74.1} &\textbf{55.1} &\textbf{49.8}  & \textbf{21.2}\\
    \midrule
    Qwen-1.5B & 41.9 &73.1 & 29.0 & 73.3 & 50.6 & 49.0 & 65.3  & 76.2 &57.3 &51.9 &56.3\\   
    +LoRA &43.3&73.9&31.2&77.6&\textbf{54.9}&48.8&66.5&76.9&59.1&53.9& 56.3\\   
    +LoRAMoE &43.9&73.7&29.8&77.3&53.4&48.7&66.3&76.9&58.8&53.2& 65.7\\
    +MixLoRA &43.4&73.8&31.8&78.2&54.6&48.9&66.4&76.5&59.2& 54.2& 141.9\\   
    \rowcolor{gray!10} +LoRA-Gen & \textbf{44.3} &\textbf{74.3} &\textbf{33.4} & \textbf{79.6} & {53.6} & \textbf{49.1} & \textbf{67.4} & \textbf{76.9} & \textbf{59.8} &\textbf{55.0}  & \textbf{26.7}\\
    \midrule
    Gemma-2B & 50.3&81.5&33.8&73.4&49.3&55.6&71.5&78.7& 61.8 &57.0& 87.3\\   
    +LoRA &49.9&78.2&36.0&\textbf{80.9}&56.8&55.4&71.7&79.2&63.5&59.2& 87.3\\ 
    +LoRAMoE &50.9&\textbf{82.0}&38.8&78.4&55.2&54.0&\textbf{72.9}&79.3&63.9&60.0&101.8\\   
    +MixLoRA &\textbf{52.3}&79.4&38.6&75.6&\textbf{59.1}&54.1&72.7&78.2&63.8&60.2&177.7\\   
    \rowcolor{gray!10} +LoRA-Gen & {51.2} &{81.9} &\textbf{39.0} & {76.2} & {55.6} & \textbf{56.0} & {71.6} & \textbf{79.5} & \textbf{63.9}& \textbf{60.2} & \textbf{36.1}  \\ 
    \bottomrule
    
    \end{tabular}
    \caption{Comparison of the performance with 5-shot samples on various commonsense reasoning benchmarks. Seen tasks indicate that the datasets are part of the training set, while unseen tasks are not. AVE denotes the average accuracy of 8 tasks while HAR is the harmonic mean. The latency scores of various methods are all calculated on ARC-c. Latency is measured on a Nvidia A100 GPU.}
    \label{tab:benchmark}
    % \vspace{-5pt}
\end{table*}
\begin{table*}[t]
    \centering
    \footnotesize
    \renewcommand\arraystretch{1.0}
    \setlength{\tabcolsep}{5.3pt}
    \begin{tabular}{l|cc|ccccc|cc}
    \toprule
    \multirow{2}{*}{Method} & \multirow{2}{*}{W/ Training}& W/ Tools & \multirow{2}{*}{$\mathrm{SR}_t$} & \multirow{2}{*}{$\mathrm{SR}_{act}$} & \multirow{2}{*}{$\mathrm{SR}_{args}$} & \multirow{2}{*}{$\mathrm{SR}$} & \multirow{2}{*}{IoU} &{Average} &{Compress} \\
    &&Definiton&&&&&&Score $\uparrow$&Ratio $\uparrow$ \\
    \midrule
    Gemma-2B & \colorxmark  & \colorcmark  & 86.3 & 77.6 & 77.7& 65.0 & 89.7& 79.3& \multirow{2}{*}{1x}\\   
    +LoRA & \colorcmark  & \colorcmark & \textbf{99.4}  & 79.6 & \textbf{93.8} & 78.2& 91.0 & 88.4 &\\   
    \rowcolor{gray!10} 
    +LoRA & \colorcmark  & \colorxmark & 98.0  &60.9 & 83.2 &52.1 & 81.3 & 75.1& \\   
    \rowcolor{gray!10} 
    +LoRA-Gen & \colorxmark  & \colorxmark & {94.1} & {86.8} & {79.7}& {73.3}& {86.9} & {84.2} & 10.1x\\ 
    \rowcolor{gray!10} 
    +LoRA-Gen & \colorcmark  & \colorxmark & {98.6} & \textbf{88.0} & {93.4}& \textbf{84.0}&\textbf{93.6} &\textbf{91.5} &\\ 
    \bottomrule
    
    \end{tabular}
    \caption{Performance of different fine-tuning strategies with Gemma-2B~\citep{gemma} on test set of GPT4Tools~\citep{gpt4tools}.
    W/ Training denotes Gemma-2B is fine-tuning on the training set of GPT4Tools with vanilla LoRA or our LoRA-Gen.
    Gray rows indicate scenarios where the system prompt does not contain tools definitions, typically constituting 91\% of the input context.}
    % Compress ratio represents the ratio of the context length.}
    \label{tab:agent}
    % \vspace{-10pt}
\end{table*}

\begin{table}[h]
    \centering
    \footnotesize
    \renewcommand\arraystretch{1.0}
    \setlength{\tabcolsep}{2.35pt}
    \begin{tabular}{l|ccc|cc}
    \toprule
    \multirow{2}{*}{Method} & \multirow{2}{*}{HellaS } & \multirow{2}{*}{WinoG } & \multirow{2}{*}{PIQA } & \multirow{2}{*}{AVE. $\uparrow$} & {Latency}      \\  &&&&& (ms)$\downarrow$ \\ 
    % Method & Type & Image Size & s & d & d &d  
    \midrule

    {AutoCompressors} & {44.7} &{62.4} &{73.3}  & {60.1} & {11.4ms}\\
    {LoRA-Gen} & \textbf{46.3} & \textbf{63.7} & \textbf{74.9} & \textbf{61.6} & \textbf{7.54ms}\\
    \bottomrule
    
    \end{tabular}
    \caption{Comparison with AutoCompressors~\cite{ChevalierWAC23} in unseen tasks based on OPT-2.7B.}
    \label{tab:autocompressors-compa}
    \vspace{-10pt}
\end{table}
\subsection{Implementation Details}
We deploy LLaMA3-8B~\citep{llama3} as the cloud-side LM during online task-specific LoRA parameters generation.
We finetune the q and v projection layers of the LLM with a LoRA adapter.
The number of experts is $8$ and we set K in the routing function TOP-K to $2$ by default.
The coefficient $\alpha$ for auxiliary loss $\mathcal{L}_{cv}$ is set 0.01.
All the latencies are measured on the same GPU with 40GB of memory.
More details can be viewed in the Appendix.
\subsection{Main results}
\paragraph{Reasoning Tasks.}
We first evaluate the performance of LoRA-Gen in the reasoning scenario as shown in~\Cref{tab:benchmark}.
We divide eight commonly used datasets into two parts, one as the multi-task learning set, including ARC-c, ARC-e, OpenBookQA, BoolQ, SocialQA and the other as an unseen test set, including Hellaswag, Winogrande and PIQA. We randomly sample to construct multi-shot training data.
% As shown in~\Cref{tab:benchmark}, LoRA-Gen consistently outperforms other fine-tuning methods across different backbone models.
As shown in~\Cref{tab:benchmark}, LoRA-Gen consistently achieves comparable performance while exhibiting lower latency compared to other fine-tuning methods across various backbone models.
Additionally, As shown in~\Cref{tab:autocompressors-compa}, based on the same LLM, our method achieves absolute gains of $1.5\%$ over AutoCompressors~\cite{ChevalierWAC23}, while maintaining much higher efficiency.
The results underscore the advantage of using LoRA-Gen, which balances effectiveness and efficiency across both seen and unseen tasks.

\paragraph{Intelligent Agent Scenario.}
We evaluate the performance of LoRA-Gen with edge-side model Gemma-2B on the GPT4Tools benchmark~\citep{gpt4tools}.
The results in~\Cref{tab:agent} present a comparison of successful rates, intersection-over-union (IoU), average performance, and compression ratio (speedup).
One key advantage of LoRA-Gen is to compress the tools definition within the system prompt into the generated LoRA parameters via a single-turn inference.
It significantly reduces the context length with a compression ratio of 10.1x, which maintains comparable performance of 91.5\% average score.
On the other hand, our method without training on GPT4Tools boosts original Gemma-2B by 4.9\% in average score, which shows the effective generalization of our method.
In contrast, removing the tool definitions in the vanilla LoRA setting leads to a marked reduction in performance ($\mathrm{SR}$: -26.1\%, IoU: -9.7\%).
Furthermore, benefiting from knowledge injection from the cloud-side language model, it surpasses the baseline by 3.1 points while maintaining a 10.1x compression ratio.
The results highlight the strengths of LoRA-Gen in effectiveness and efficiency, attributed to its inference-time specialization and generalization ability to unseen tools, making it well-suited for tasks with extensive prefix descriptions.
\begin{figure*}[t]
    \centering
    \includegraphics[width=0.95\linewidth]{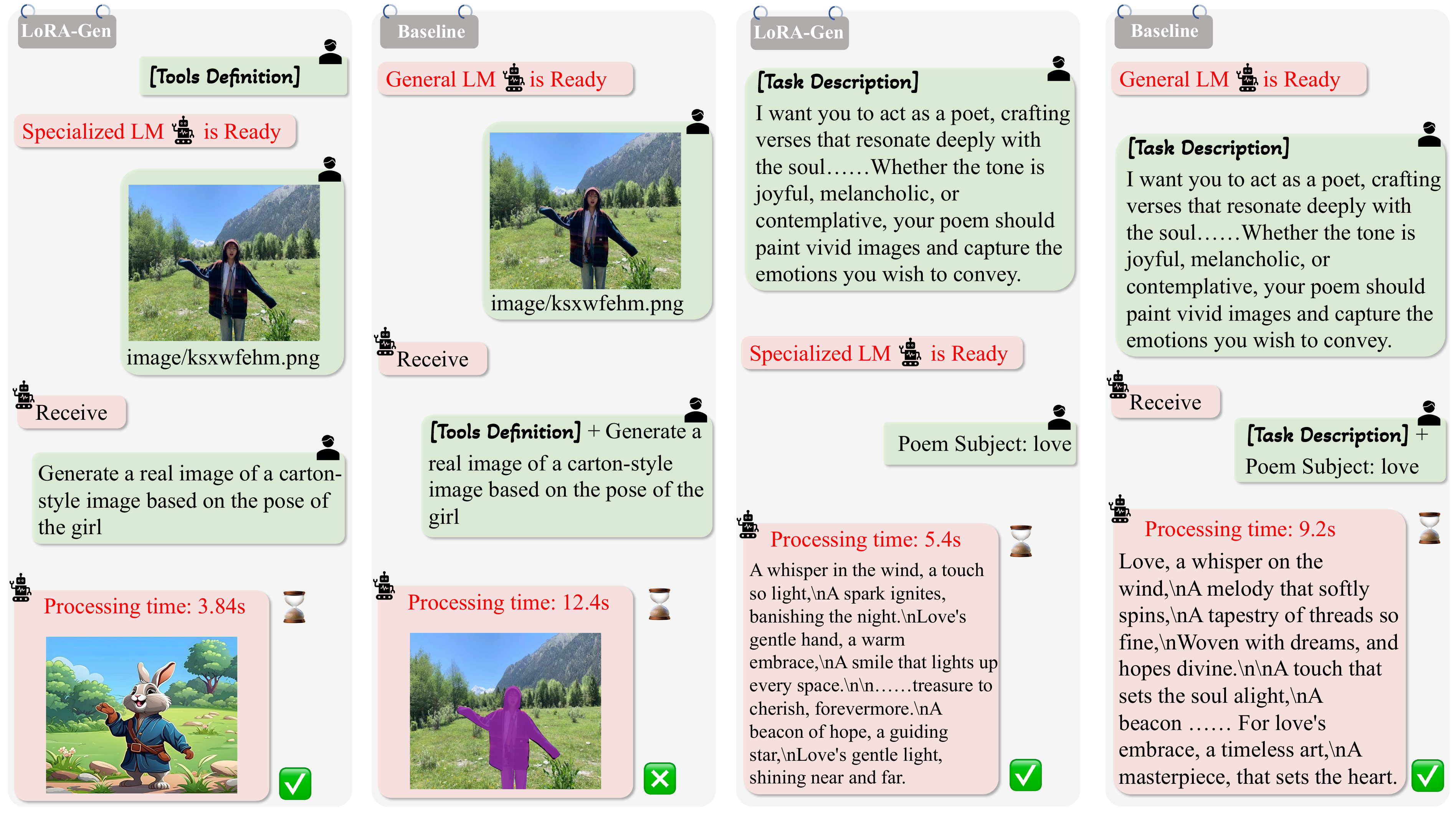}
    \caption{Visualization comparison between LoRA-Gen and baseline, Gemma-2B~\citep{gemma}.
    LoRA-Gen compresses the tools definition and task description into the generated LoRA parameters, effectively specializing the language model to reduce processing times while maintaining comparable performance.
    The detailed LM outputs and system prompt can be accessed in the Appendix.}
    \label{fig:case}
\end{figure*}
\subsection{Ablation study}
\label{sec:ablation}
\paragraph{Number of Experts in Online Expert Pool.}
As shown in~\Cref{tab:number_expert}, we present the performance of different numbers of experts in the cloud-side LoRA pool. 
Performance generally improves with an increasing number of experts.
With 4 experts, the AVE. is 56.4\%, and the HAR. is 52.3 \%.
Increasing the number of experts to 12 yields slight improvements, with the AVE. rising to 57.3\% and the HAR. to 53.1\%.
However, the best performance is achieved with 8 experts, where both AVE. (58.7\%) and HAR. (53.6\%) reach their peak values.
This may indicate that 8 experts strike the best balance between multi-task learning and unseen generalization.
\paragraph{Effectiveness of Balanced Load Strategy.}
Ensuring a balanced load of experts can significantly improve the robustness and stability of the model.
We initially conduct an ablation study to assess the impact of the absence of auxiliary losses on model performance.
Without the auxiliary loss, the AVE. decreases by 1.2 points.
Subsequently, we summarize the impact of different values of the coefficient for auxiliary loss as shown in~\Cref{tab:coef}.
As the auxiliary loss coefficient decreases, a significant improvement in both performance metrics is observed.
Reducing the coefficient from 0.1 to 0.01 yields further gains, resulting in an average (AVE) of 58.7\% and a harmonic mean (HAR) of 53.6\%, thereby achieving an optimal balance between the auxiliary strategy and the primary objective function.
In addition, we investigate the strategy of the router function.
As illustrated in \Cref{tab:router}, we compare two routing strategies employed for online experts within the cloud-side LoRA pool. 
Compared to Gumbel-softmax, KeepTOP-K strategy exhibits a notable improvement, attaining an AVE of 58. 7\% and a HAR of 53. 6\%.
We consider that an overabundance of randomness may affect expert ability to learn specific tasks during the optimization process.
\input{table/ablation}
\input{table/ablation_2}
\paragraph{Effectiveness of Meta Token.}
We attempt to utilize the cloud-side large language model to generate LoRA parameters in a single forward pass directly instead of meta tokens.
Specifically, we directly transform the output tokens of LLM to the LoRA weights space with a feedforward neural network and get the $i$-th layer generated LoRA weights $\in  \mathbb{R}^{3\times2\times d \times r}$, where $d$ is the hidden dimension and $r$ denotes the low rank of LoRA.
As indicated by the experimental results in~\Cref{tab:lora_generate}, this approach exhibits comparable performance to that achieved through meta tokens on the seen tasks, while the results on the unseen tasks are significantly lower than those obtained with meta tokens, trailing by 11.1\%.
Generating LoRA parameters directly leads to pronounced overfitting to the training domain, caused by the large parameter space, thereby limiting its ability to generalize to unseen tasks.
\paragraph{Effectiveness of Knowledge Transfer.}
As depicted in~\Cref{tab:few-shot}, we compare the performance of the baseline model and our LoRA-Gen across different few-shot samples.
Remarkably, LoRA-Gen with just a 1-shot sample surpasses the baseline with 5-shot samples by 3.5\% on HAR.
We attribute this to the use of LLaMA3-8B~\citep{llama3} as the cloud model, which transfers a portion of its knowledge to the edge-side language model via reparameterization.

\subsection{Qualitative Study in Agent Scenario}
We deploy LoRA-Gen within Gemma-2B and conduct case studies and visualizations.
As illustrated in~\Cref{fig:case}, LoRA-Gen removes the 26 tools description from input of the model, significantly reducing inference time and achieving a 3.2x speedup compared to the baseline.
The limited generalization of the baseline model results in incorrect tool selection, thereby highlighting the effectiveness of our method.
Additionally, in the open text generation scenario, LoRA-Gen accelerates reasoning time by compressing the task definition while achieving comparable results.
The corresponding generation results are detailed in the appendix.

\vspace{-0.2cm}
\section{Conclusion}
In this paper, we propose an online LoRA generation framework, called LoRA-Gen, which utilizes a cloud-side language model to generate task-specific LoRA parameters for edge-side models.
Our strategy offers four advantages over previous methods: context compression for unseen tasks, a reparameterized language model, inference-time specialization, and knowledge transfer.
Extensive experiments show that LoRA-Gen achieves competitive results and an impressive speedup on common-sense reasoning tasks.
Additionally, our method achieves a compression ratio of 10.1x on zero-shot agent tasks, indicating its potential applicability to more scenarios. We believe our methodological approach can inspire future LLM-based research.
\section*{Acknowledgement}
This work was partly supported by Shenzhen Key Laboratory of next generation interactive media innovative technology (No:ZDSYS20210623092001004) and  National Natural Science Foundation of China (No.62293544, 62425117).
\section*{Impact Statement}
This paper presents work whose goal is to advance the field of 
Machine Learning. There are many potential societal consequences 
of our work, none which we feel must be specifically highlighted here.

\bibliography{icml2025}
\bibliographystyle{icml2025}

\newpage
\section{Appendix}
\subsection{Training details}
The models are trained with eight NPUs (64GB memory per device) by default.
We set betas and momentum of the AdamW optimizer with (0.9, 0.999) and 0.9, respectively.
During training, we utilize a Cosine Scheduler with an initial learning rate of $2\times10^{-5}$ and weight decay of 0.1.
The details are shown in~\Cref{tab:exp_cfg}
\begin{table}[h]
\centering
\renewcommand\arraystretch{1.2} % line space
\begin{tabular}{l|cc}
\toprule
Hyper-parameters          & LoRA-Gen     \\ 
\midrule
optimizer                & AdamW                             \\
learning rate            & 2e-5                  \\
warm steps               & 50                     \\
weight decay             & 0.1                          \\
optimizer momentum       & $\beta_1$, $\beta_2$=0.9, 0.999  \\
batch size               & 64                                 \\
epoch                    & 4                               \\
max length               & 2048                        \\
LoRA attention dimension (r)                     & 16                      \\
LoRA scaling alpha ($\alpha$)                      & 16                      \\
LoRA drop out                         & 0.05                    \\ 
\bottomrule
\end{tabular}
\caption{Fine-tuning configuration.}
\label{tab:exp_cfg}
% \vspace{-20pt}
\end{table}
\subsection{Detailed Assistant Output}
The definition of tools follows GPT4Tools~\cite{gpt4tools}, encompassing vision foundation models~\cite{mambatree}, generative models~\cite{sdxl}, and application-specific models~\cite{instructpix2pix, uvcom}.
Task description for the role play in the qualitative study of the main text can be seen in~\Cref{tab:role_play}.
To strengthen LoRA-Gen's ability to compress and process instructions in the system prompt, we modify the Alpaca dataset, using GPT-4 to generalize specific problems into instruction sets, which are subsequently used as training data.
\subsection{Statistical Significance}
\begin{table*}[h]
    \centering
    \footnotesize
    \renewcommand\arraystretch{1}
    \setlength{\tabcolsep}{6pt}
    \begin{tabular}{l|cccccccc}
    \toprule
    \multirow{2}{*}{Method} & \multirow{2}{*}{ARC-c  }  & \multirow{2}{*}{ARC-e}      & \multirow{2}{*}{OBQA } & \multirow{2}{*}{BoolQ }  & \multirow{2}{*}{SIQA } & \multirow{2}{*}{HellaS } & \multirow{2}{*}{WinoG } & \multirow{2}{*}{PIQA }      \\  &&&&&&&& \\ 
    % Method & Type & Image Size & s & d & d &d  
    
    \midrule
    
    {TinyLLaMA} & {0.0146} &{0.0089} &{0.0219}  & 0.0076 & {0.0112} & {0.0050} & {0.0134} & {0.0100}\\
    {Qwen} & {0.0145} &{0.0089} &{0.0229}  & 0.0071 & {0.0113} & {0.0050} & {0.0132} & {0.0098}\\
    {Gemma} & {0.0146} &{0.0089} &{0.0218}  & 0.0075 & {0.0112} & {0.0050} & {0.0135} & {0.0096}\\
    \bottomrule
    
    \end{tabular}
    \caption{\textbf{Standard error on language model benchmarks.}.}
    \label{tab:std-error}
    \vspace{-10pt}
\end{table*}
The standard errors of different tasks are shown in~\Cref{tab:std-error}, all statistics are calculated with the open-sourced lm-evaluation-harness project~\citep{eval-harness}. Additionally, we have re-evaluated our method 4 times on GPT4Tools with a variation of about 0.65\% in average score.

\subsection{Training Data.}
\Cref{tab:data-size} outlines the data scale for each reasoning task. Moreover, we process the Alpaca dataset through GPT-4, resulting in a filtered and abstracted set of 37,658 training samples.
\begin{table*}[ht]
    \centering
    \footnotesize
    \renewcommand\arraystretch{1}
    \setlength{\tabcolsep}{6pt}
    \begin{tabular}{l|cccccccc}
    \toprule
    \multirow{2}{*}{Method} & \multirow{2}{*}{ARC-c  }  & \multirow{2}{*}{ARC-e}      & \multirow{2}{*}{OBQA } & \multirow{2}{*}{BoolQ }  & \multirow{2}{*}{SIQA } & \multirow{2}{*}{HellaS } & \multirow{2}{*}{WinoG } & \multirow{2}{*}{PIQA }      \\  &&&&&&&& \\ 
    % Method & Type & Image Size & s & d & d &d  
    
    \midrule
    
    {Train} & {1120} &{2250} &{4957}  & 9427 & {33410} & {39905} & {9248} & {16100}\\
    {Test} & {1171} &{2380} &{500}  & 3270 & {1954} & {10042} & {1267} & {1838} \\

    \bottomrule
    
    \end{tabular}
    \caption{\textbf{The data size of tasks used in our experiments.}}
    \label{tab:data-size}
    % \vspace{-10pt}
\end{table*}

\subsection{Efficiency Comparison}
\begin{table*}[h]
    \centering
    \footnotesize
    \renewcommand\arraystretch{1.0}
    \setlength{\tabcolsep}{4.2pt}
    % \begin{tabular}{l | c | c c c | c c c}
    \begin{tabular}{l|ccc|ccc}
    \toprule
    \multirow{2}{*}{Method} & \multicolumn{3}{c}{Training Mode}  & \multicolumn{3}{c}{Inference Mode} \\
    \cline{2-7} 
    &  FLOPs & Memory & Latency & FLOPs & Memory & Latency   \\ 
    \midrule
    +LoRA &4.736E+11 &37096MiB & 0.85s&4.708E+11 & 11208MiB & 0.19s\\
    +LoRAMoE &4.742E+11 &26326MiB &1.19s &4.742E+11 &11286MiB & 0.22s\\   
    +MixLoRA$^\dagger$ &5.061E+11 &30844MiB & 2.17s & 5.048E+11   & 11828MiB & 1.08s\\   
     +LoRA-Gen & 1.667E+12 &39603MiB &2.84s & 1.552E+11 & 10932MiB & 0.11s\\
    \bottomrule
    
    \end{tabular}
    \caption{\textbf{Efficiency Comparison.}}
    \label{tab:efficient_comparison}
    % \vspace{-5pt}
\end{table*}
\Cref{tab:efficient_comparison} presents the efficiency Comparison among different approaches. MixLoRA$^\dagger$ indicates the method without specific optimization. All metrics are measured on an Nvidia GPU. FLOPs are measured using an input of 100 tokens and an instruction of 200 tokens, while memory and latency are evaluated in training mode with a batch size of 8 per GPU.

\begin{table*}[]
\centering
\begin{tcolorbox} 
\renewcommand\arraystretch{1.2} % line space
\begin{tabular}{p{\textwidth}}
\textbf{\textcolor{blue}{Task description of role play:}}

I want you to act as a poet, crafting verses that resonate deeply with the soul. Your task is to create a poem that not only evokes emotions but also possesses the power to move and inspire those who read it. The subject matter will be provided by the user, and your words should be carefully chosen to express the essence of that theme in a way that touches hearts and stirs deep feelings. Aim for beauty in your language—let your choice of words carry both meaning and emotion, so that each line speaks to the reader on a profound level. Whether the tone is joyful, melancholic, or contemplative, your poem should paint vivid images and capture the emotions you wish to convey.

\\
\textbf{\textcolor{blue}{Edge-side model with LoRA-Gen answer:}}

A whisper in the wind, a touch so light,$\backslash$A spark ignites, banishing the night.$\backslash$Love's gentle hand, a warm embrace,$\backslash$A smile that lights up every space.$\backslash$$\backslash$Two souls entwined, a tapestry of dreams,$\backslash$Where laughter dances and love softly gleams.$\backslash$A bond unbreakable, a trust so deep,$\backslash$In each other's eyes, secrets they keep.$\backslash$$\backslash$Through storms and sunshine, hand in hand they roam,$\backslash$Building castles of hope, a love that calls them home.$\backslash$A silent language, spoken in the heart,$\backslash$A symphony of feelings, a work of art.$\backslash$$\backslash$For love is a journey, a path to explore,$\backslash$A treasure to cherish, forevermore.$\backslash$A beacon of hope, a guiding star,$\backslash$Love's gentle light, shining near and far. 

\\
\textbf{\textcolor{blue}{The baseline answer:}}

Love, a whisper on the wind,$\backslash$A melody that softly spins,$\backslash$A tapestry of threads so fine,$\backslash$Woven with dreams, and hopes divine.$\backslash$$\backslash$A touch that sets the soul alight,$\backslash$A beacon in the darkest night,$\backslash$A warmth that melts the coldest fear,$\backslash$A solace, ever drawing near.$\backslash$$\backslash$It blooms in laughter, shared and bright,$\backslash$In silent moments, bathed in light,$\backslash$A gentle hand, a knowing glance,$\backslash$A shared embrace, a sacred dance.$\backslash$$\backslash$Love's strength endures, a steadfast guide,$\backslash$Through storms and calm, it cannot hide,$\backslash$A constant echo, soft and deep,$\backslash$A promise whispered, secrets to keep.$\backslash$$\backslash$So let us cherish every breath,$\backslash$This precious gift, beyond all death,$\backslash$For love's embrace, a timeless art,$\backslash$A masterpiece, that sets the heart. $\backslash$$\backslash$$\backslash$Let me know if you'd like me to explore a specific aspect of love, or if you have a particular style or tone in mind.

\end{tabular}
\end{tcolorbox}
\caption{Detailed supplement to the visualization results in the main text.}
\label{tab:role_play}
\end{table*}

\subsection{More Potential Application}
Our current validation focuses on LLMs, and future work will explore its applicability to multimodal large models~\cite{haplo, qwenvl} and unified systems~\cite{mindomni, transfusion, haploomni} for understanding and generation.
In addition, we plan to enrich more models as tools by incorporating advanced image generation and editing models~\cite{flux,taiming}, as well as speech recognition and synthesis models~\cite{audioldm}.

\end{document}